\documentclass[letterpaper, 10 pt, conference]{ieeeconf}  

\IEEEoverridecommandlockouts                              

\usepackage{graphics} 
\usepackage{epsfig} 
\usepackage{amssymb}  
\usepackage{bm, amsmath,color,soulutf8,longtable,colortbl,setspace,xspace,url,pdflscape,cite}

\usepackage[font=small, labelfont=bf]{caption} 
\usepackage[font=small, labelfont=bf]{subcaption}
\usepackage{graphicx}
\usepackage{todonotes}

\title{\LARGE \bf
Real-Time Simultaneous Localization and Mapping with LiDAR Intensity
}

\author{Wenqiang Du and Giovanni Beltrame
  \thanks{The authors are with the Department of Computer Engineering and
    Software Engineering, Polytechnique Montreal, University of Montreal,
    Montreal, QC, H3T1J4, Canada. {\tt\small \{wenqiang.du,
      giovanni.beltrame\}@polymtl.ca}}%
}

\begin{document}

\maketitle
\thispagestyle{empty}
\pagestyle{empty}

\begin{abstract}
  We propose a novel real-time LiDAR intensity image-based simultaneous
  localization and mapping method, which addresses the geometry degeneracy
  problem in unstructured environments. Traditional LiDAR-based front-end
  odometry mostly relies on geometric features such as points, lines and planes.
  A lack of these features in the environment can lead to the failure of the
  entire odometry system. To avoid this problem, we extract feature points from
  the LiDAR-generated point cloud that match features identified in LiDAR
  intensity images. We then use the extracted feature points to perform scan
  registration and estimate the robot ego-movement. For the back-end, we jointly
  optimize the distance between the corresponding feature points, and the point
  to plane distance for planes identified in the map. In addition, we use the
  features extracted from intensity images to detect loop closure candidates
  from previous scans and perform pose graph optimization. Our experiments show
  that our method can run in real time with high accuracy and works well with
  illumination changes, low-texture, and unstructured environments.  
\end{abstract}

\section{INTRODUCTION}
Simultaneous Localization and Mapping (SLAM) is a fundamental problem in
robotics. SLAM is the process of building a map of the environment and in the
meanwhile, tracking the robot's pose on the generated map. There are many kinds
of SLAM methods, such as visual SLAM\cite{qin2018vins}, Light Detection and Ranging (LiDAR) SLAM\cite{zhang2014loam},
visual and LiDAR SLAM\cite{shan2021lvi}, visual-intertial and ranging SLAM (VIR-SLAM)\cite{cao2021vir} and so on.
In this paper, we focus on the LiDAR SLAM.

Numerous SLAM methods based on LiDAR\cite{shan2021lvi,liosam2020shan,qin2019aloam,shan2018lego} have been proposed over the past few years. Most of them are based on the geometric features (e.g.,
edges and planes \cite{zhang2014loam, shan2018lego}) of LiDAR point clouds.
These methods are robust and accurate in most structured environment. However,
when the environment has little structure, these methods can suffer from
geometric degeneracy and fail\cite{ebadi2021dare, xu2022fast}. For example, in a long corridor
or a cave environment, there might be enough plane features, but too few edge
features to estimate the relative ego-motion movement between two consecutive
scans. Unfortunately, {aligned} plane features {(like in the corridor)} alone are not sufficient to estimate the
robot pose in six degrees of freedom (6DOF), meaning the robot could lose the
sense of forward or backward movement.
\begin{figure}
        \centering
                \begin{subfigure}[b]{0.48\textwidth}
                \includegraphics[width=1\linewidth]{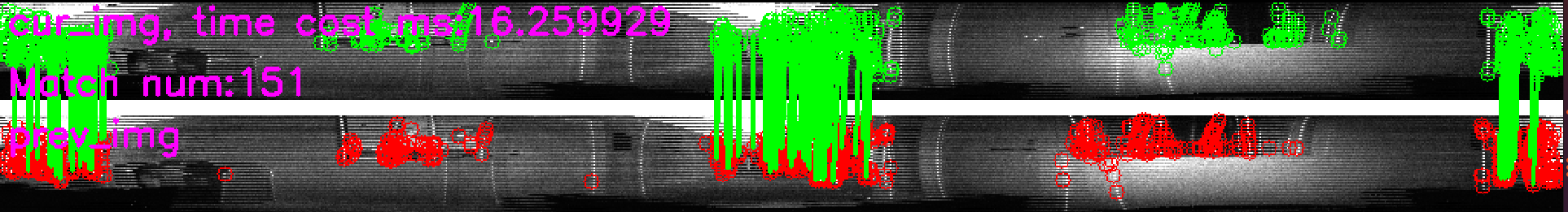}
                \caption{Feature points in the current scan (top) and the previous scan
                  (bottom) are matched with ORB features from intensity images.}
                \label{fig:intensityImg}
                \end{subfigure}
                \begin{subfigure}[b]{0.48\textwidth}
                \includegraphics[width=1\linewidth]{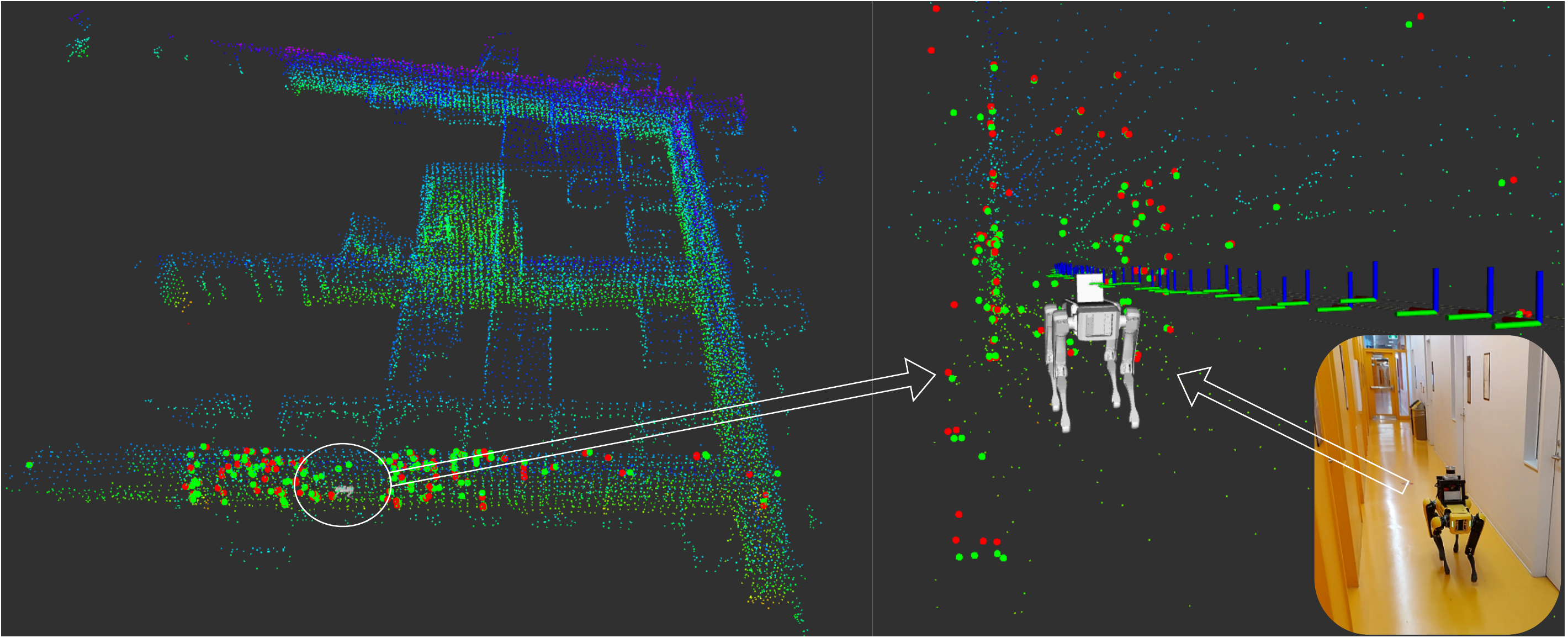}
                \caption{The matched feature points are used to perform the scan registration.}
                \label{fig:matchedPoints}
                \end{subfigure}
                \caption{The matched 3D points from two consecutive scans and
                  their corresponding feature points. The points in {\bf{(b)}}
                  are the 3D points that are extracted from point clouds
                  according to the indexes of matched features from {\bf(a)}. The red points
                  represent matched points of the previous frame, and the green
                  points stand for matched points in the current scan. Those
                  points are then used for scan registration to estimate the
                  relative poses between two consecutive
                  frames.} 
\end{figure}

To solve this problem, one needs an additional reference to constrain the sixth
degree of freedom. One possible way is to extract feature points from textural
information. Recently, many researchers tried to add intensity information to
augment \cite{li2022intensity} or assist \cite{wang2021intensity} LiDAR SLAM
systems, improving their performance. However, these feature points from
textural information are already presented in 3D space and theoretically are
enough to constrain the 6DOF movement. Why not directly perform scan
registration with these feature points? We propose a pure LiDAR intensity SLAM method
which directly extracts feature points from intensity images and perform scan
registrations to estimate the robot's ego-movement. Our contributions are:
\begin{itemize}
\item A novel real-time LiDAR intensity image-based SLAM system, aiming at
  solving the geometric degeneracy problem; 
\item Combining the benefits of visual SLAM systems {with those of} LiDAR SLAM systems,
  without suffering from blurring or illumination changes;
\item A lightweight front-end due to fewer feature points and adding
  ground plane constraint and LiDAR Bundle Adjustment (BA) to the back-end;
\item Intensity-based loop closure detection and pose graph
  optimization;
\end{itemize}

\section{RELATED WORK} \label{sec:related}
In the past decades, many researchers worked in SLAM and achieved many great
results. In 1990, Smith et al.\cite{smith1990estimating} firstly used {extended Kalman
filter} (EKF) to estimate the relative position among objects in an unknown
environment and build a stochastic map to store these estimations of spatial
relationships and their uncertainties.
Nowadays, there are many kinds of SLAM systems~\cite{huang2019survey}, many
using an Inertial measurement unit (IMU), cameras, and/or LiDARs.

LiDAR-based SLAM systems have been well studied in both theory and industrial
applications in recent years.
One representative work is LOAM~\cite{zhang2014loam}, a real-time method for
estimating odometry and mapping. LOAM only consumes a LiDAR's point cloud and
then extracts edge and planar points from this point cloud to register
consecutive scans at 10Hz and providing map updates at 1Hz. Another popular work
is Cartographer~\cite{hess2016real}, which uses a scan-to-submap matching
strategy with loop closure detection and graph optimization, achieving high
accuracy even in real-time. LeGO-LOAM~\cite{shan2018lego} is also a real-time
odometry and mapping SLAM method, which uses only the LiDAR of ground vehicles
as a front-end sensor. LeGO-LOAM extracts ground planar and edge features and
assigns them different labels, which significantly decreases the number of
features. Shang et al.~\cite{liosam2020shan} proposed LIO-SAM, which is a
tightly-coupled LiDAR-inertial odometry system. LIO-SAM can estimate the
odometry by using LiDAR and a 9-axis high-frequency IMU. Shang et al. results
showed that LIO-SAM is much more accurate than LOAM and can be run in real-time
on a computationally limited platform.

Differentiating from these feature-based odometry methods, Chen et
al.~\cite{chen2022direct} proposed a direct method to estimate the odometry
using LiDAR and IMU. Their results showed that their method is more accurate
than a feature-based approach. DLO~\cite{chen2022direct} is another direct LiDAR
odometry algorithm that can match consecutive frames of point cloud directly
with high accuracy in real-time for computationally-limited robot platforms, but
it is sensitive to dynamic objects. Lin et
al.~\cite{lin2019fast}\cite{lin2020loam} developed a robust and accurate SLAM
system using solid-state LiDAR with small field of view (FoV).

In addition to traditional algorithms, some deep learning-based LiDAR SLAM
{~\cite{chen2019iros, chen2022overlapnet, cattaneo2022lcdnet}}
systems also achieved good performance. PointNet~\cite{qi2017pointnet} and PointNet++~\cite{qi2017pointnet++} leverage deep learning techniques to directly extract semantic features from 3D point clouds, enabling their use in segmentation and extraction of semantic information. In addition to
feature extraction, deep learning was used in SegMatch~\cite{dube2017segmatch}
for place recognition matching 3D segments.
PointNetVLAD~\cite{uy2018pointnetvlad} is a combination of PointNet and
NetVLAD~\cite{arandjelovic2016netvlad}, which uses end-to-end learning to
extract global features from a frame of point cloud, which are very useful for
place recognition.

Recently, as LiDAR resolution is steadily increasing, we can generate much more
clear and textured images (see Fig.~\ref{fig:intensityImg}) from intensity
information. {In recent years, integrating intensity information into their SLAM system has emerged as a novel approach among researchers~\cite{barfoot2016into,shan2021robust, guadagnino2022fast}.} Wang et al.~\cite{wang2021intensity} introduce intensity features
into their SLAM system, using both intensity and geometric features to improve
the performance of the SLAM system. Li et al.~\cite{li2022intensity} extract
intensity edge points in a solid-state LiDAR-inertial SLAM system, while~\cite{di2021visual, shan2021robust} proposed a novel visual place recognition method using LiDAR intensity information.

In this paper, we propose a novel lightweight LiDAR odometry method which
directly matches 3D feature points extracted from intensity images. This method
can merge both the feature tracking ability of visual SLAM and LiDAR's high accuracy.
We also propose an intensity image-based back-end, including additional constraints
between intensity features and using intensity information to detect
loop closures. 

\begin{figure*}[h]
        \centering
        \includegraphics[width=1\linewidth]{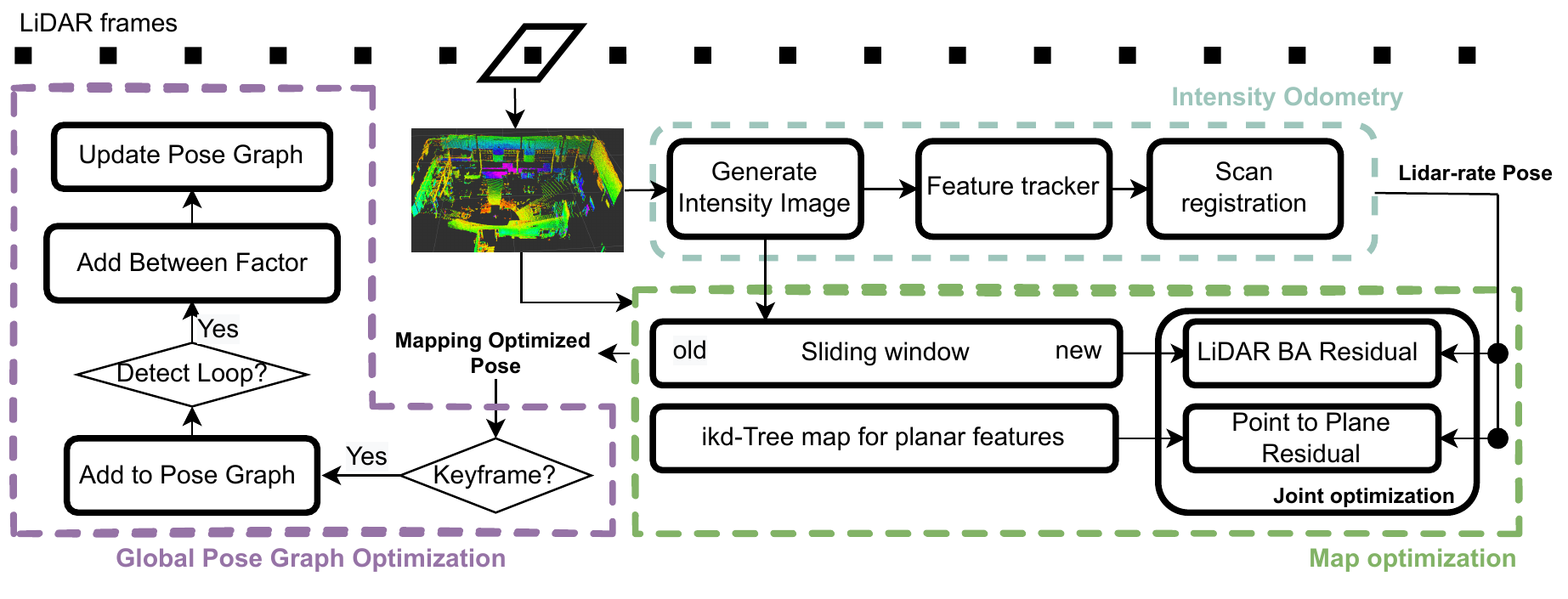}
        \caption{System overview of the proposed method. The whole system
          consists of three parts, including intensity odometry, map
          optimization, and pose graph optimization. The intensity odometry part
          is the core of the proposed method. It consists of intensity image
          generation, feature tracking, and scan registration. The map
          optimization corrects the drift by jointly minimizing both LiDAR BA
          residual and point to map plane residual. Pose graph optimization
          corrects the whole trajectory by adding loop constraints.}
        \label{fig:flow}
\end{figure*}
\section{METHOD} \label{sec:method}

The pipeline of our proposed method is shown in Fig.~\ref{fig:flow}. In our
system, a LiDAR generates a point cloud within 100ms that is called a frame or a
scan. To estimate the movement of the robot, we need to calculate the relative
pose between consecutive frames. We use intensity odometry to implement this
procedure in the front-end. In addition, we use the term ``odometry'' to
describe the relative pose between the current frame and the initial frame. The
odometry from the front-end is generally inaccurate, so we need to use a
back-end to optimize the odometry. In the back-end, we use the scan-to-map
method\cite{zhang2014loam} and LiDAR BA\cite{liu2021balm} to correct the drift. However, map
optimization is generally not enough for removing the accumulated drift, and we
add loop closure detection as an additional means to reduce drift. In our case,
we perform loop closure detection on the LiDAR intensity image and use pose
graph optimization to update the trajectory and generate the final trajectory of
the robot.

\subsection{Intensity Odometry}
Assume that we have two consecutive frames of point clouds $\mathcal{X}=
\{\mathbf{X}_1, \mathbf{X}_2, \cdots, \mathbf{X}_J \}$ and $ \mathcal{Y} =
\{\mathbf{Y}_1, \mathbf{Y}_2, \cdots, \mathbf{Y}_I \}$ from a LiDAR: one direct
way of estimating the relative pose is directly applying an Iterative Closest
Point (ICP) algorithm\cite{rusinkiewicz2001efficient} to calculate the rotation matrix
$\mathbf{R}$ and $\mathbf{T}$:

\begin{align}\label{eq:icp}
	\mathop{\arg\min}_{\mathbf{R,T}} \sum\limits_{\mathbf{X}_j\in \mathcal{X}, \mathbf{Y}_i \in \mathcal{Y}}\parallel \mathbf{Y}_i - \mathbf{R}\mathbf{X}_j - \mathbf{T} \parallel^2
\end{align}

However, this method usually consumes significant time and computation
resources\cite{zhang2014loam}. To reduce the computation cost, we need to extract
representative points for the scan registration, thereby reducing the number of
points used for optimization (up to a limit: we still need to maintain the
original relationships between two frames).

In order to decrease the number of points used for optimization, ~Zhang et al.\cite{zhang2014loam} tried to extract edge and plane features. The points of the edge feature of the current frame can be matched with the edges in the map. The same goes for plane features.

With edge and plane features, we can optimize the point to line distance,
and point to plane distance jointly, and estimate $\mathbf{R}$ and
$\mathbf{T}$. However, sometimes we cannot extract edge features accurately enough, like in
the long corridor or cave environment. In this case, we will lose the ability to
estimate 6DOF movement.

To solve this issue, we extract and track features directly from intensity
images. Fig.~\ref{fig:intensityImg} shows the intensity images generated from an
Ouster-64 LiDAR, where the image resolution is $1024 \times 64$. Even if the
vertical resolution is low, we can still extract enough features (the red and
green circles are the ORB features\cite{mur2015orb, mur2017orb,campos2021orb}) for estimating movement. We
extract 3D points, $\mathcal{Y}_2 = \{\mathbf{Y}_1, \mathbf{Y}_2, \cdots, \mathbf{Y}_k \}$
directly from the point cloud according to the index of the matched {ORB} feature
points from the intensity image. Each 3D feature point is assigned a score $\mathcal{S}
\in \{\mathbf{S}_1, \mathbf{S}_2, \cdots, \mathbf{S}_n\}$ obtained during feature extraction. Similarly, we
can extract the corresponding points $\mathcal{X}_2 = \{\mathbf{X}_1, \mathbf{X}_2,
\cdots, \mathbf{X}_k \}$ in the following scan, and we scan match as a least
squared estimation problem: 
\begin{align}
	\label{eq:xn} \bar{\mathbf{X}}_n &= \mathbf{RX}_n + \mathbf{T} \\
	\label{eq:argmin_2} \mathop{\arg\min}_{\mathbf{R,T}} &\sum\limits_{n\in \mathcal{N}} \parallel(\mathbf{Y}_n - \bar{\mathbf{X}}_n)\cdot \mathbf{S}_n \parallel^2
\end{align}
where $\mathcal{N}=[1, 2, \cdots, k]$, and $\mathbf{S}_n$ is the score of the $n$th
matched feature point {according to Hamming distance}. This way, we can limit the number of features to around
200 points.

\subsection{Map Optimization}
The LiDAR intensity odometry generates a transformation matrix
\begin{equation*}
\mathbf{\hat{T}} =
        \begin{bmatrix}
         \mathbf{R} & \mathbf{T} \\
                \mathbf{0} & 1
        \end{bmatrix}
\end{equation*}        
between current sensor frame and the map frame. In this module, we jointly optimize the {scan-to-map} residual and LiDAR BA residual to correct the pose drift.

\subsubsection*{LiDAR Bundle Adjustment (BA)}
\begin{figure}
        \centering
        \includegraphics[width=1\linewidth]{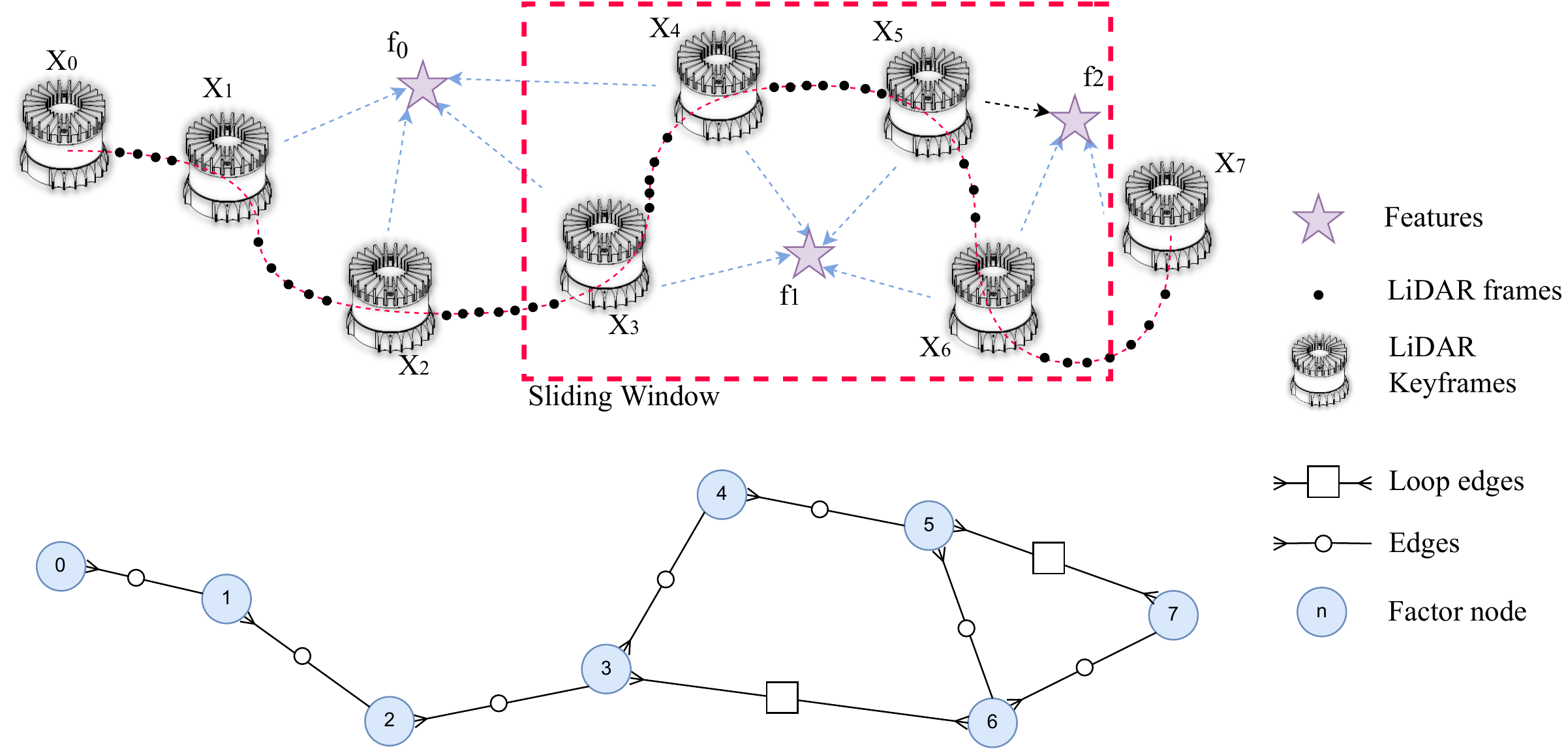}
        \caption{Illustration of the sliding window strategy used for LiDAR BA (top) and Pose graph (bottom) with loop closure contraints.}
        \label{fig:pgo_sliding_window}
\end{figure}

\begin{figure*}[]
        \centering
	\begin{subfigure}{0.19\textwidth}
		\includegraphics[width=\linewidth]{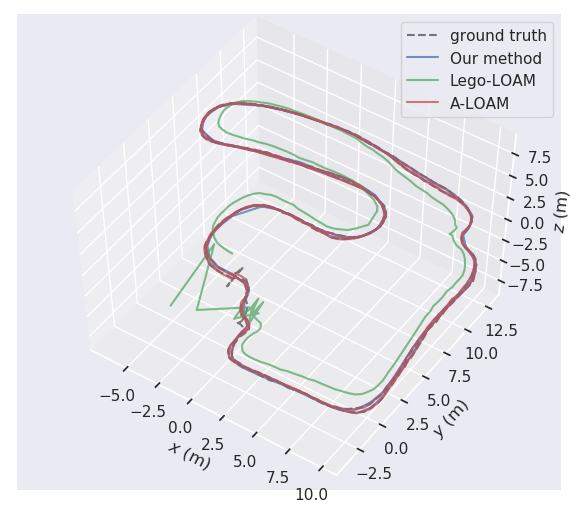}
		\caption{Trajectories in outdoor environment with up and down stairs (191m).}
		\label{fig:traj_indoor1}		
	\end{subfigure}
	\begin{subfigure}{0.19\textwidth}
		\includegraphics[width=\linewidth]{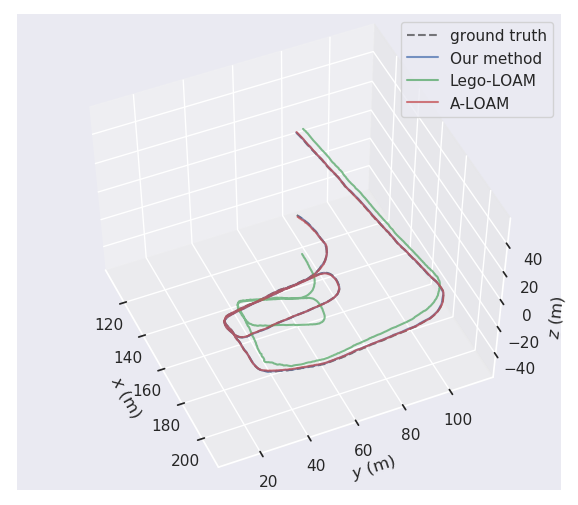}
		\caption{Trajectories in outdoor environment with steep slope (414m).}
		\label{fig:handheld3}		
	\end{subfigure}
	\begin{subfigure}{0.19\textwidth}
		\includegraphics[width=\linewidth]{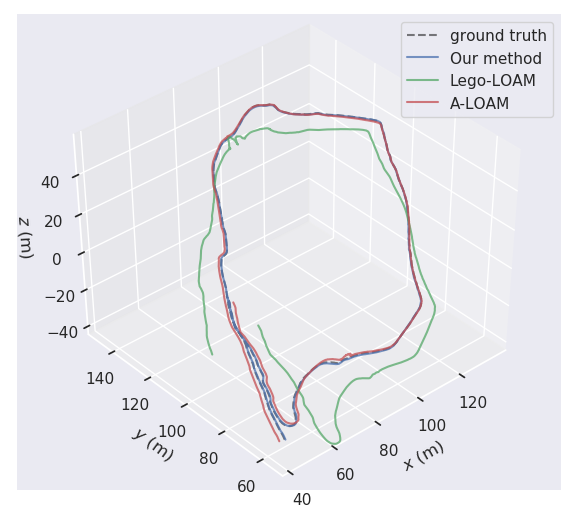}
		\caption{Trajectories on mountain with a long slope (507m).}
		\label{fig:jackal1}		
	\end{subfigure}
        \begin{subfigure}{0.19\textwidth}
		\includegraphics[width=\linewidth]{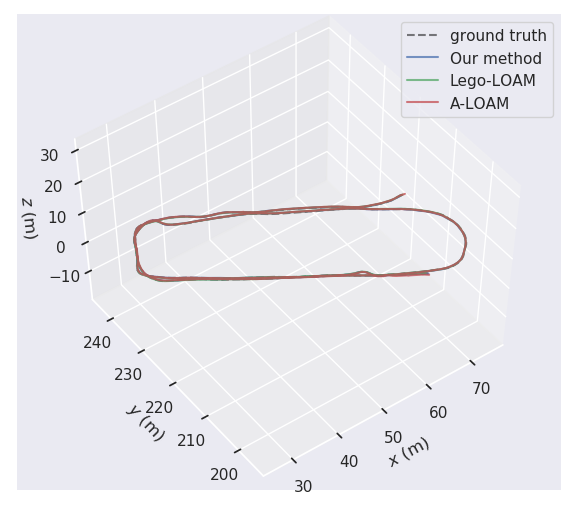}
		\caption{Trajectories in parking lot with flat ground (249m).}
		\label{fig:traj_jackal3}		
	\end{subfigure}
        \begin{subfigure}{0.19\textwidth}
		\includegraphics[width=\linewidth]{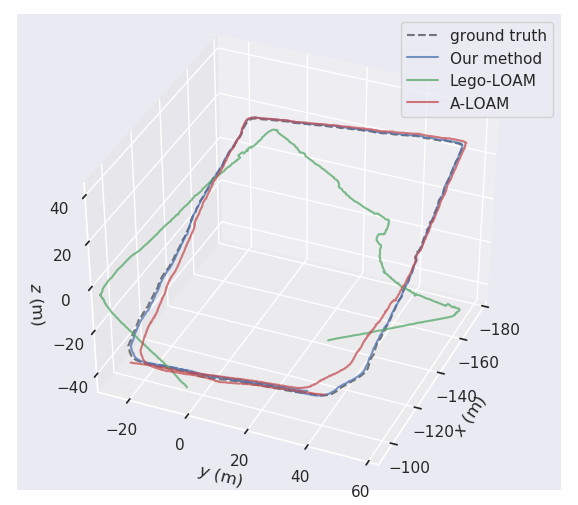}
		\caption{Trajectories on the street with revisiting the start point (478m).}
		\label{fig:handheld2}		
	\end{subfigure}
	\caption{Trajectories results in multiple environments. The experimental results prove that our method is able to estimate the position accurately in various scenarios. LeGO-LOAM's algorithm works well in flat environments, but not in environments with slopes.}
	\label{fig:trajtories_outdoor}
\end{figure*}

\begin{figure*}[]
        \centering
	\begin{subfigure}{0.19\textwidth}
		\includegraphics[width=\linewidth]{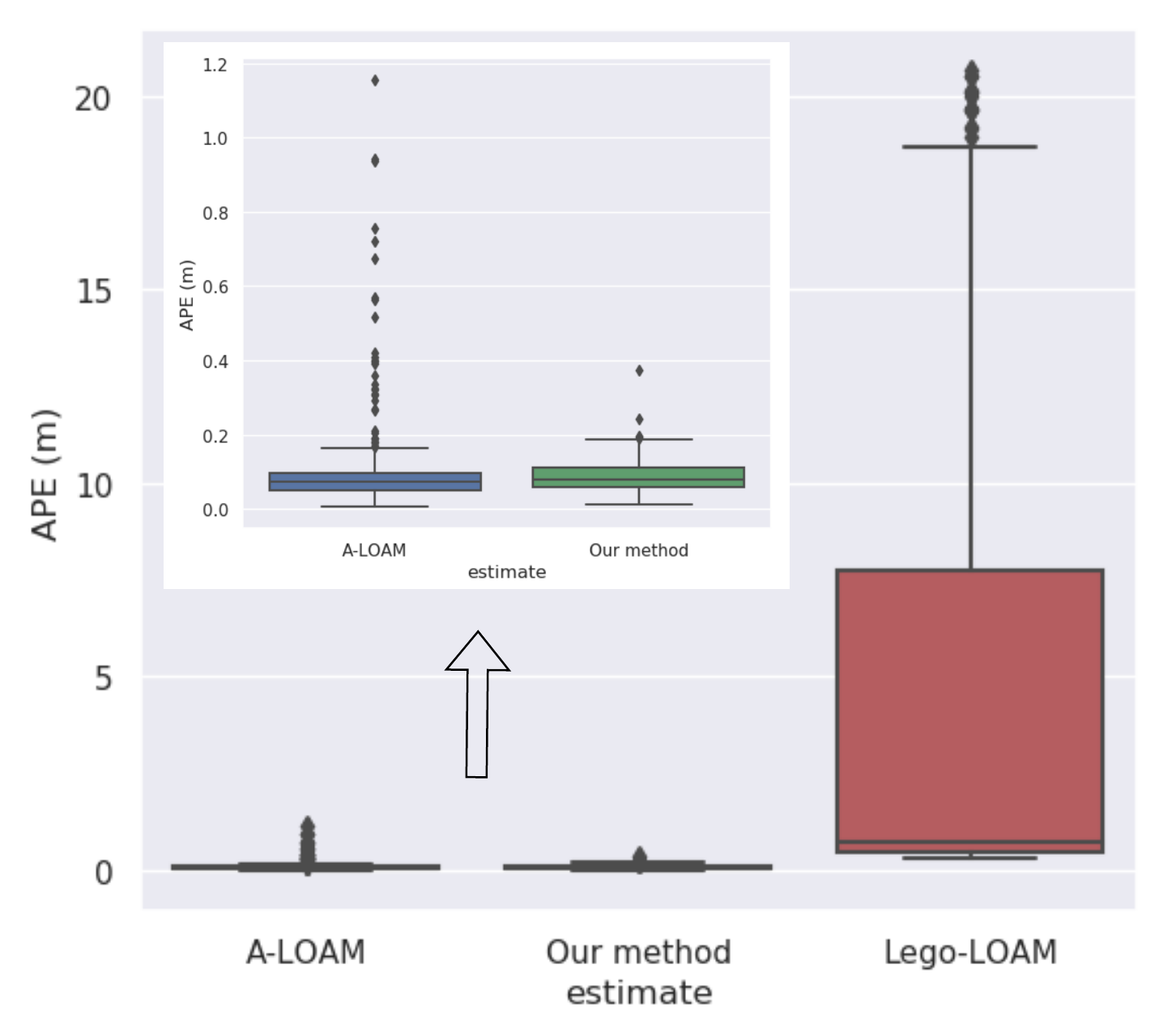}
		\caption{}
		\label{fig:traj_indoor1_ape_box}		
	\end{subfigure}
	\begin{subfigure}{0.19\textwidth}
		\includegraphics[width=\linewidth]{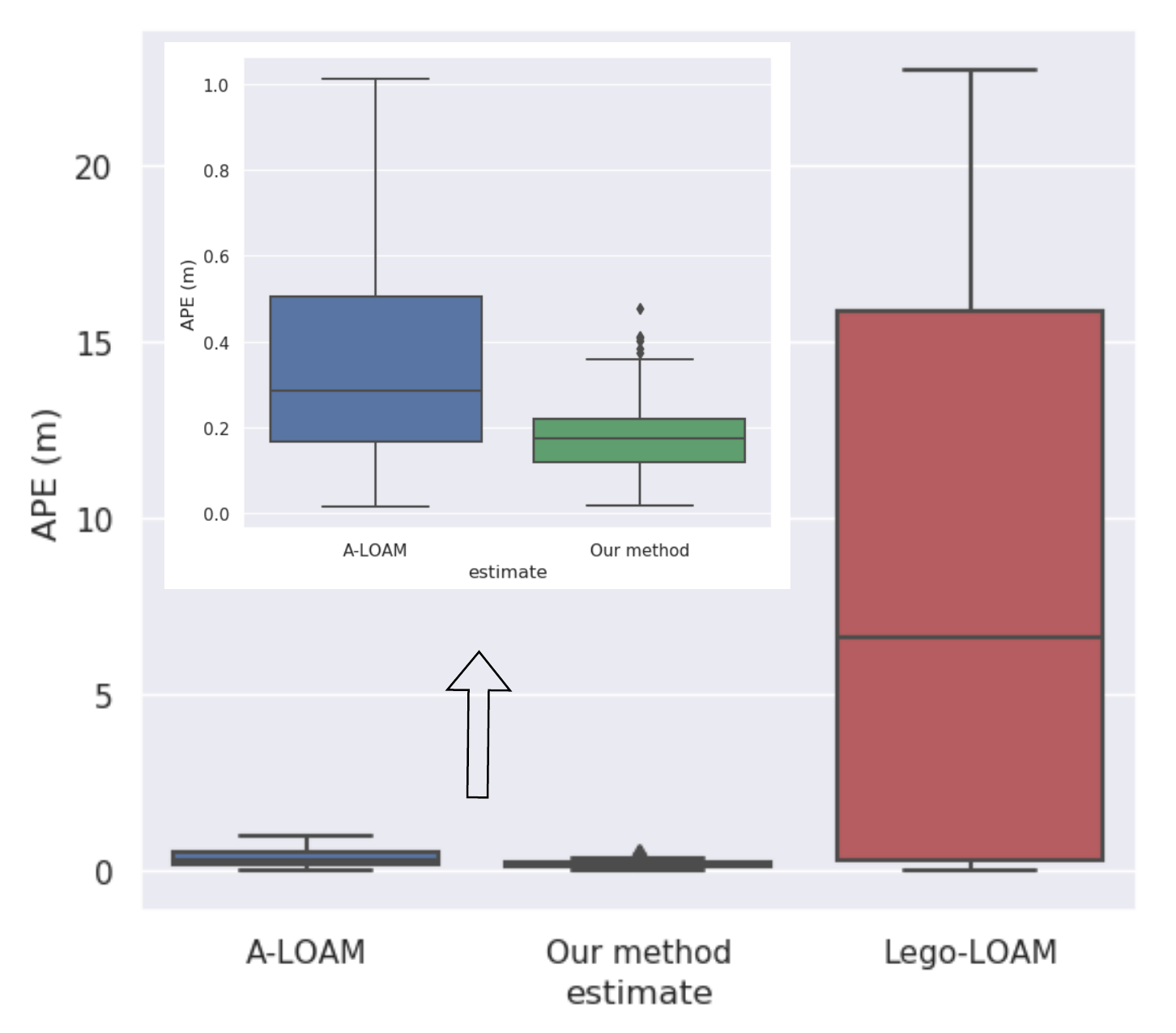}
		\caption{ }
		\label{fig:handheld3_ape_box}		
	\end{subfigure}
	\begin{subfigure}{0.19\textwidth}
		\includegraphics[width=\linewidth]{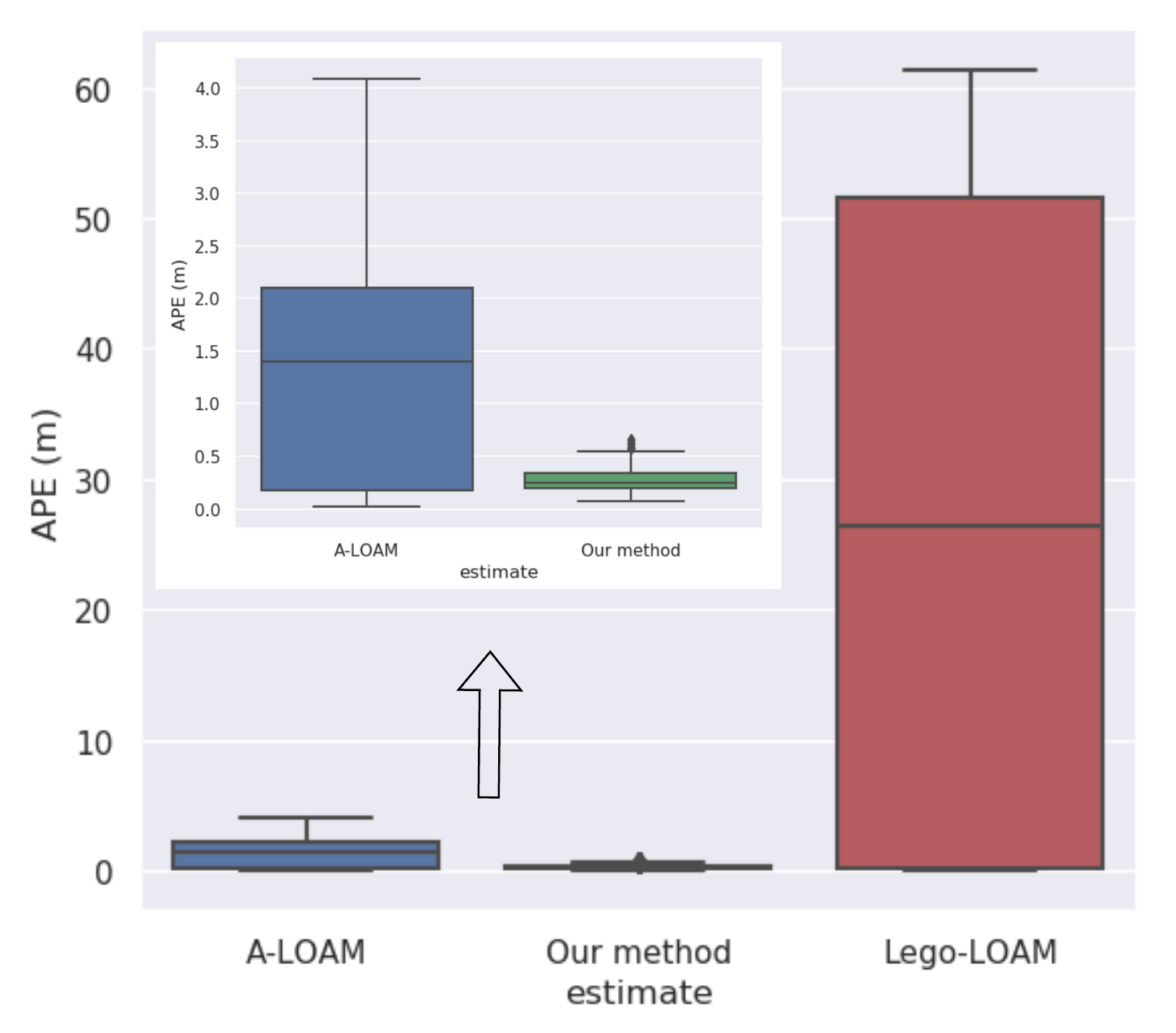}
		\caption{}
		\label{fig:jackal1_ape_box}		
	\end{subfigure}
        \begin{subfigure}{0.19\textwidth}
		\includegraphics[width=\linewidth]{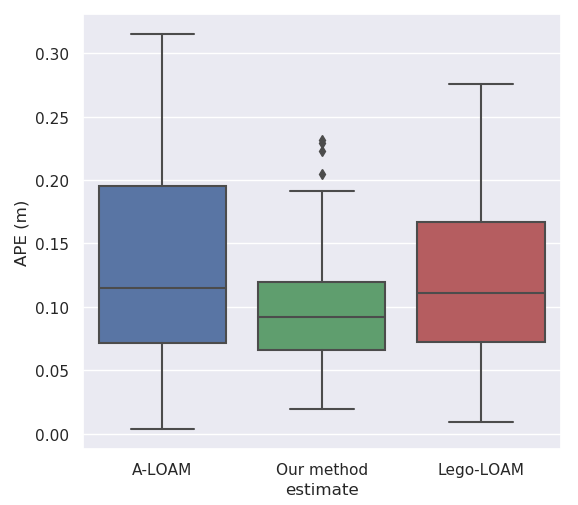}
		\caption{}
		\label{fig:traj_jackal3_ape_box}		
	\end{subfigure}
        \begin{subfigure}{0.19\textwidth}
		\includegraphics[width=\linewidth]{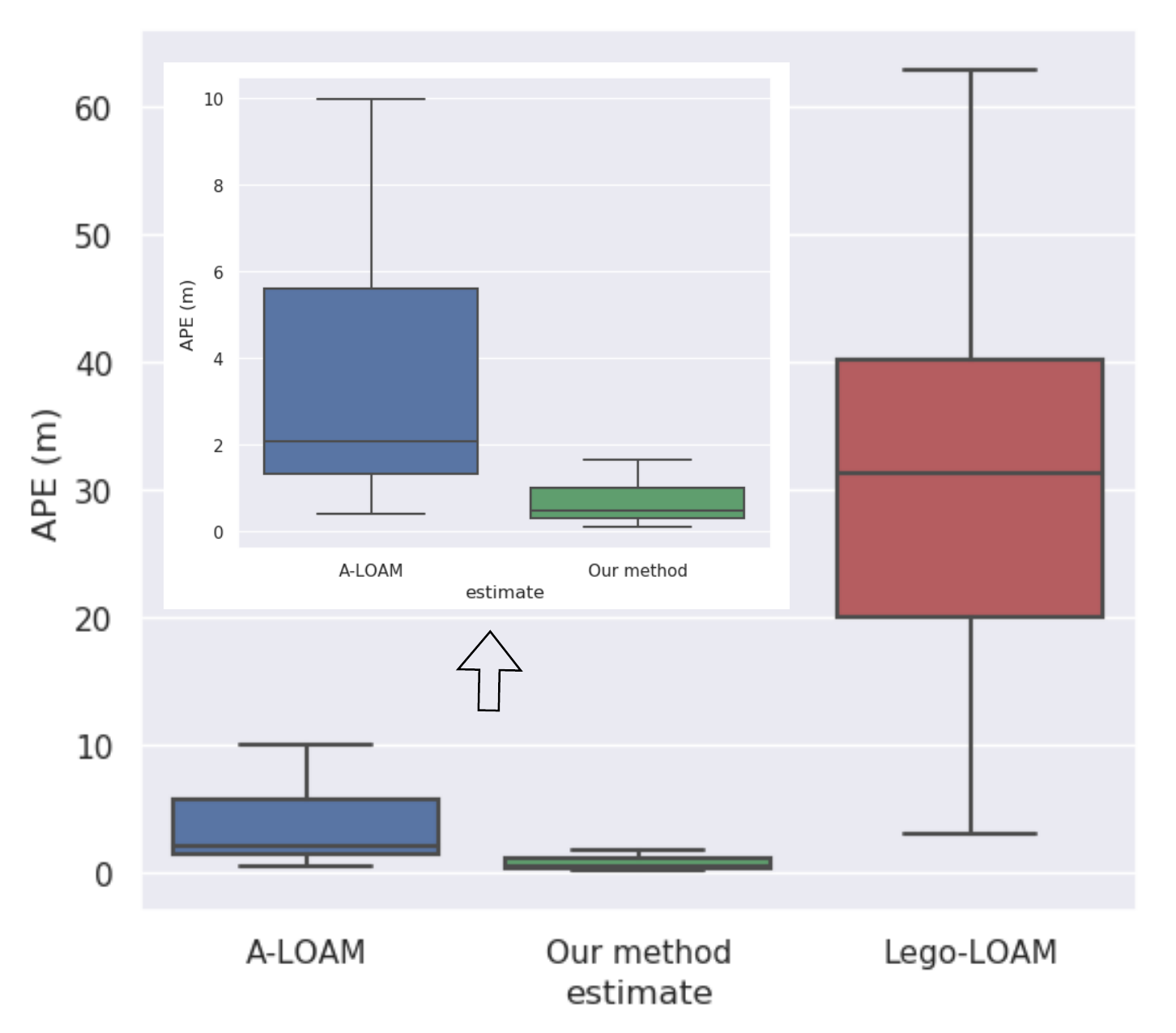}
		\caption{}
		\label{fig:handheld_part2_ape_box}		
	\end{subfigure}
	\caption{Absolute Pose Error in multiple environments. The APE of our method is the smallest in all these scenarios, followed by A-LOAM.}
	\label{fig:trajtories_outdoor_ape}
\end{figure*}

Similar to visual SLAM BA, we can use the LiDAR BA (a non-linear optimization
problem) to correct the drift. With this strategy, the last $k$ frames are used
in a residual function. We remove the oldest frame if the number of frames
is larger than $k$, and add the newest frame to the sliding window as Fig.~\ref{fig:pgo_sliding_window} showed. 
We can then match the current frame with the last $k$ frames in the
window and store the matched 3D feature points in $\mathcal{P}^c=\{\mathbf{P}^c_0,
\mathbf{P}^c_1, \cdots, \mathbf{P}^c_k \}$, and $\mathcal{F}^l=\{ \mathbf{F}^l_0, \mathbf{F}^l_1,
\cdots, \mathbf{F}^l_k\}$ for current and last $k$ frames, respectively. Where $\mathbf{P}^c_i = \{ \mathbf{P}^c_{i0}, \mathbf{P}^c_{i1}, \cdots, \mathbf{P}^c_{im} \}$,
$\mathbf{P}^c_{ij}=[{p_x^{ij}, p_y^{ij}, p_z^{ij}, 1}]^{T}$,  The
transformation matrix of the last $k$ frames are calculated and stored in
$\mathcal{\hat{T}}^l= \{ \mathbf{\hat{T}}_0, \mathbf{\hat{T}}_1, \cdots, \mathbf{\hat{T}}_k \}$ by
the previous map optimization step. The residual function is therefore:
\begin{align}
        \label{eq:lidarBa} 
        \bm{f}_b = \sum\limits_{i=0}^k \sum\limits_{j=0}^m (\mathbf{\hat{T}}\mathbf{P}^c_{ij} - \mathbf{\hat{T}}_i\mathbf{F}^l_{ij} )
\end{align}
where $\bm{f}_b$ is the residual of LiDAR BA. In addition, we treat $\mathbf{\hat{T}}$
from the intensity odometry as an initial pose of current frame.

\subsubsection*{Scan-to-Map}
To match with previous plane feature points, we create an
ikd-Tree\cite{cai2021ikd} map, which is a incremental 3D k-d Tree. This map
incrementally updates a k-d tree with incoming plane feature points. Compared to
static k-d trees, this method has lower computation cost. For plane features, we
divide the plane into ground plane and general plane features. When meeting a flat area, adding ground plane constraints is better than plane only method since the ground plane can constrain the roll and pitch direction more accurately. We extract normal plane features like in~\cite{zhang2014loam}. As for the ground plane, we first give an initial height
of robot to extract possible ground planes and then use RANSAC to segment the
ground plane points. The plane features are extracted from the point cloud, and
are stored in $\mathcal{P}^m=\{\mathbf{P}^m_0, \mathbf{P}^m_1, \cdots,
\mathbf{P}^m_n \}$, where $\mathbf{P}^m_0=\{{p_x, p_y, p_z}\}^{T}$. For each
plane feature point $\mathbf{P}_i^p$, we can find 3 nearby plane points from the ikd-Tree map, and
store them in $\mathcal{F}^m=\{ \mathbf{F}^m_0, \mathbf{F}^m_1, \cdots,
\mathbf{F}^m_n\}$, where $ \mathbf{F}^m_i=\{ \mathbf{P}_{i}^{m0},
\mathbf{P}_{i}^{m1}, \mathbf{P}_{i}^{m2} \}$. The residual function can be
formulated as:
\begin{equation}
        \label{eq:pointConvert} 
        \mathbf{\hat{P}}_i^m = \mathbf{R}\mathbf{P}_i^m + \mathbf{T}
\end{equation}
\begin{equation}
        \label{eq:point2plane}
        \bm{f}_s = \sum\limits_{i=0}^n (\mathbf{\hat{P}}_i^m - \mathbf{P}_i^{m0})^T \cdot 
                \left(\frac{(\mathbf{P}_i^{m1} - \mathbf{P}_i^{m0})\times (\mathbf{P}_i^{m2} - \mathbf{P}_i^{m0})}{|  (\mathbf{P}_i^{m1} - \mathbf{P}_i^{m0})\times (\mathbf{P}_i^{m2} - \mathbf{P}_i^{m0}) |}\right)
\end{equation}
where $\mathbf{\hat{P}}_i^m$ is the 3D mapped point which was
converted by $\mathbf{P}_i^m$ from the sensor to the map coordinate system through
\eqref{eq:pointConvert}. $\mathbf{P}_i^{m0}$, $\mathbf{P}_i^{m1}$, and $\mathbf{P}_i^{m2}$
are nearby points of $\mathbf{\hat{P}}_i^m$ on the map.

After getting the residual function of LiDAR BA and scan-to-map, we can formulate
the whole optimization problem as: 
\begin{equation}
        \label{eq:jointOptimization}
        \mathop{\arg\min}_{\mathbf{R,T}}( \bm{f}_b + \bm{f}_s)
\end{equation}
      
\begin{figure}
        \centering
        \begin{subfigure}{0.23\textwidth}
                \includegraphics[width=\linewidth]{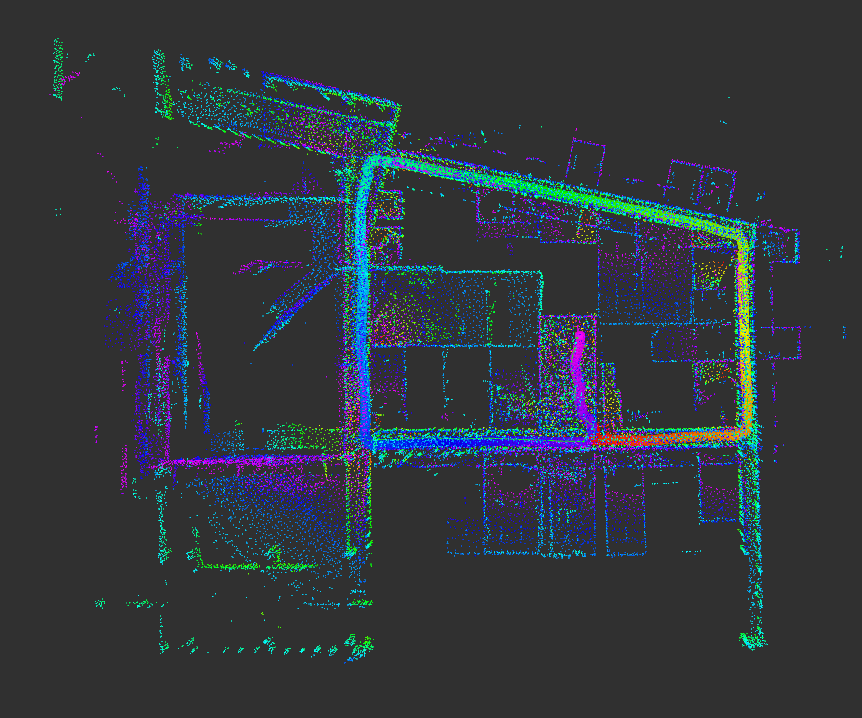}
                \caption{Map generated by our method in the long corridor environment.}
                \label{fig:spot_corridor1_map}
        \end{subfigure}
        \begin{subfigure}{0.23\textwidth}
                \includegraphics[width=\linewidth]{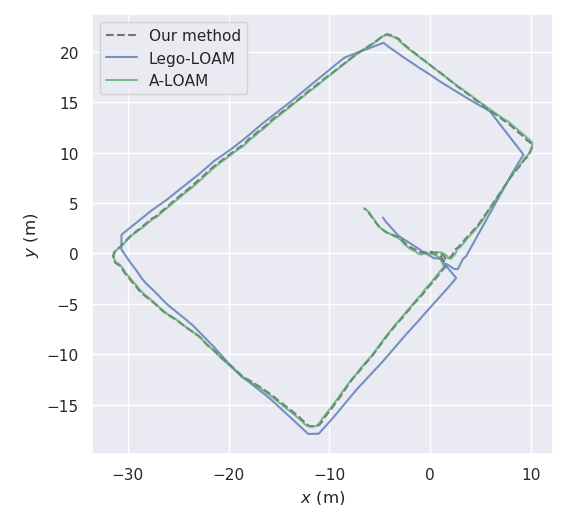}
                \caption{Trajectories of A-LOAM, LeGO-LOAM and our method.}
                \label{fig:spot_corridor1_traj_xy}
        \end{subfigure}  
        \caption{Map and trajectories of the Spot robot in a building with long corridors. In this scene, we walked along the corridor back to the starting point. In this experiment, the drift of LeGO-LOAM is relatively large, and the performance of A-LOAM is close to our method, but it is not clear from these two graphs which method is better.}
        \label{fig:spot_corridor1}    
\end{figure}
\begin{figure}
        \centering
        \begin{subfigure}{0.23\textwidth}
                \includegraphics[width=\linewidth, height=3.5cm]{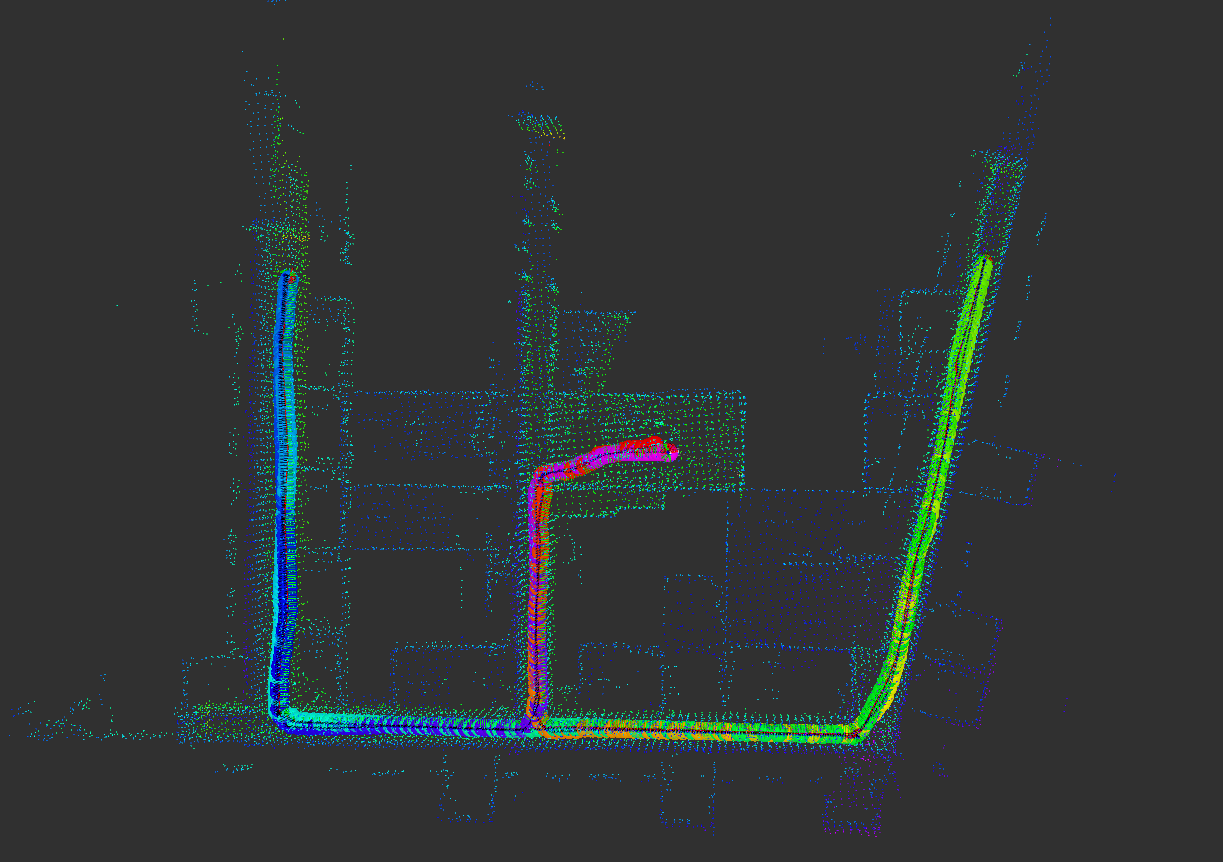}
                \caption{Map result of our method with a different path.}
                \label{fig:spot_corridor2_map}
        \end{subfigure}
        \begin{subfigure}{0.23\textwidth}
                \includegraphics[width=\linewidth, height=3.5cm]{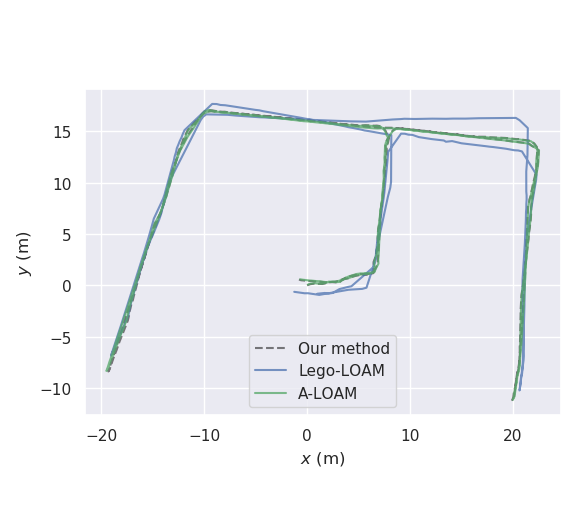}
                \caption{Trajectories of different methods. }
                \label{fig:spot_corridor2_traj}
        \end{subfigure}
        \caption{Instead of going back around in a circle to the starting point, we went back and forth through all the corridors for this test and then returned to the starting point. This situation is extreme because there is no good loop to correct the drift in the pitch direction. The experimental results show that LeGO-LOAM drifts more than in the previous scenario. A-LOAM and our method are almost the same.}
        \label{fig:trajtories_spot2}
\end{figure}

\subsection{Pose Graph Optimization}
During map optimization, we can get a better pose estimation of the current frame. Once done, we can use optimized results to correct the drift for the
future frames and publish high-frequency optimized odometry in real time. In
the backend, we build a pose graph on top of map optimization with LiDAR
keyframes. First of all, we extract keyframes from whole LiDAR frames with three
criteria:
\begin{itemize}
\item The distance between current frame and last keyframe is larger than a
  threshold.
\item The angle between two keyframes is larger than a threshold.
\item The number of matched feature points is less than a threshold.
\end{itemize}
We can use the optimized poses of keyframes as the vertexes of the pose graph, and the
relative poses between two keyframes as the edges of pose graph.

We also add loop constraints to the pose graph, as Fig.~\ref{fig:pgo_sliding_window}
shows, we use the latest keyframe as the anchor frame. With a trained
vocabulary, we can compare descriptors of the current keyframe with a database where the history
descriptors are stored. If we can not match it with history descriptors, then no
loop closure is found for this keyframe~\cite{shan2021robust}. If we
successfully match a previous keyframe, we can then put it into the outlier
rejection procedure to test whether it is an incorrect loop closure. If it is
positive, we can add this loop constraint between the current and
the loop candidate factor node in the pose graph. Finally, we use
g2o\cite{kummerle2011g} to solve the pose graph optimization problem.

\section{EXPERIMENTAL RESULTS} \label{sec:experiment}
\begin{figure}
        \centering
        \begin{subfigure}{0.3\textwidth}
                \includegraphics[width=0.8\linewidth, height=4cm]{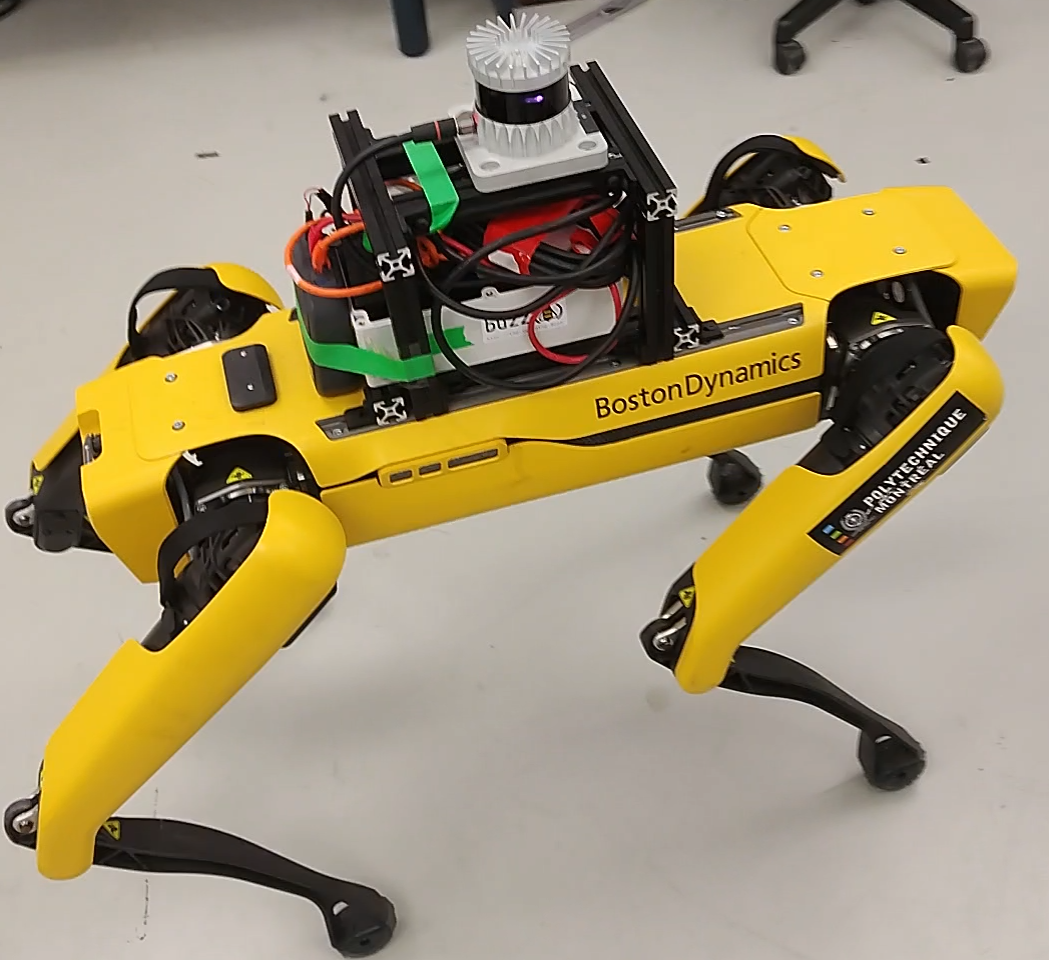}
                \caption{Spot robot for indoor experiment}
                \label{fig:spot_equipment}
        \end{subfigure}
        \begin{subfigure}{0.17\textwidth}
                \includegraphics[width=0.9\linewidth, height=4cm]{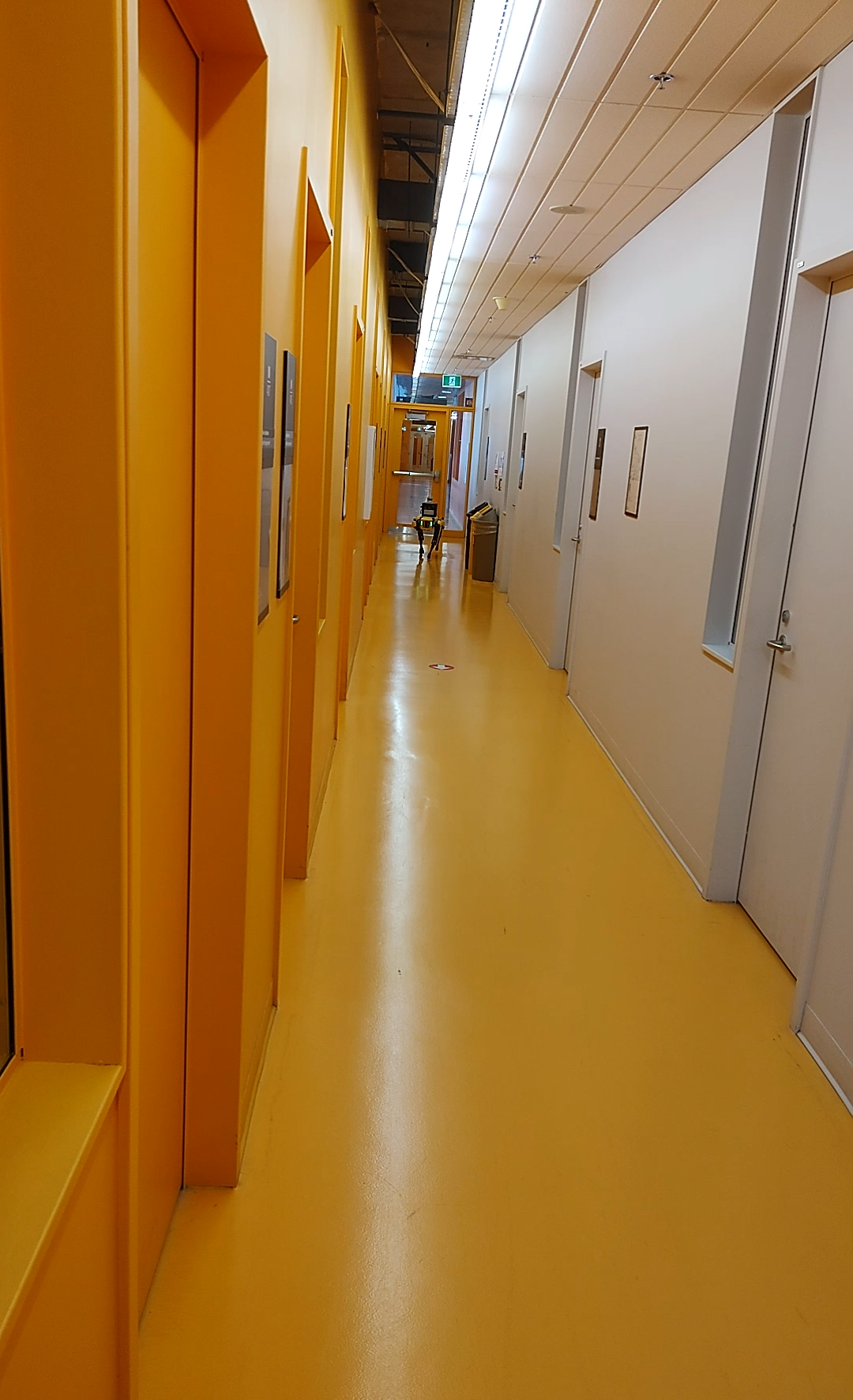}
                \caption{Indoor corridor}
                \label{fig:experiment_env_robot_environment}
        \end{subfigure}
        
        \caption{Indoor environment and Spot robot used for testing our algorithm. Spot with Ouster LIDAR mounted on it walking and collecting data in the corridor shown on the right.}
        \label{fig:experiment_env_robot}        
\end{figure}

To prove our algorithm's reliability, we present our experimental results in an
indoor environment with long corridors (Fig.~\ref{fig:experiment_env_robot_environment}), a multi-storey indoor environment, a
mountain, and a street environment. The reason why we chose these environment is that they are all different from each other, and they are challenging for pure LiDAR SLAM. In the indoor environment, we have long corridors and narrow passages. In the mountain environment, we have steep slopes and narrow passages. In the street environment, we have many obstacles and many turns. Those scenarios are difficult for pure LiDAR SLAM to handle. In the following, we will introduce the details of our experiments.

\begin{figure}
        \centering
        \begin{subfigure}{0.48\textwidth}
                \includegraphics[width=\linewidth]{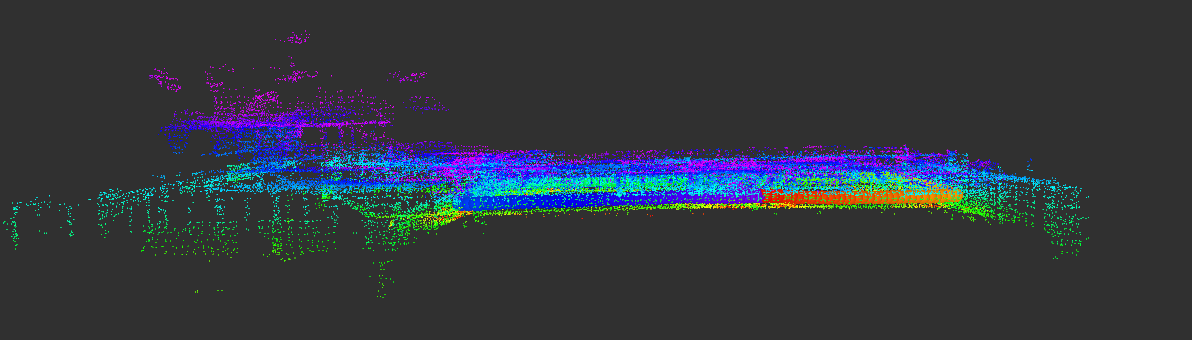}
                \caption{The front view of the map of our method}
                \label{fig:spot_corridor1_intensity_map_details}
        \end{subfigure}
        \begin{subfigure}{0.48\textwidth}
                \includegraphics[width=\linewidth]{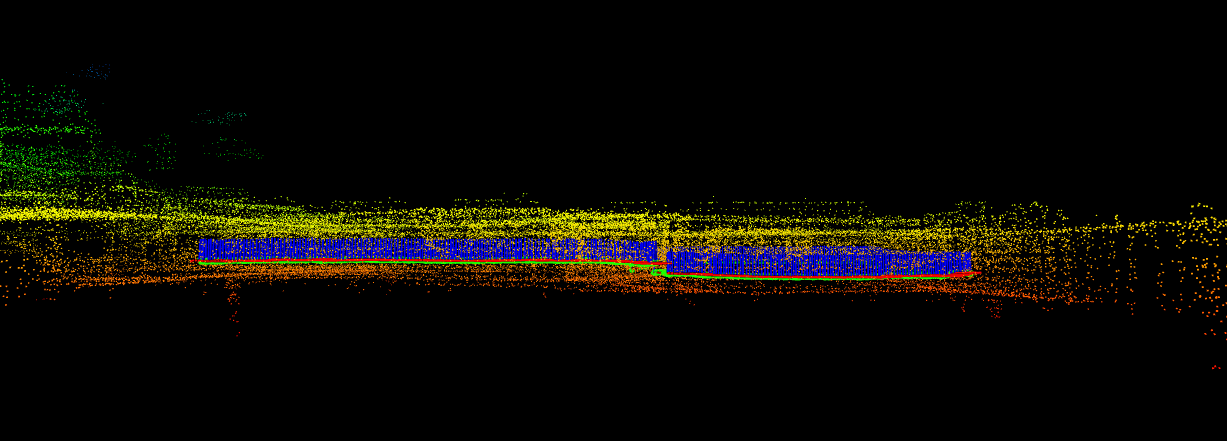}
                \caption{The front view of the map of A-LOAM}
                \label{fig:spot_corridor1_aloam_map_details}
        \end{subfigure}
        \caption{Difference between our method and A-LOAM in long corridors. As we can see, the map generated by our method is smoothly connected to the starting position, but the map generated by A-LOAM has a clear break at the end.}
        \label{fig:diff_aloamandour}
\end{figure}
   
In our experiments, we compared our approach
with two other popular pure LiDAR SLAM systems: LeGO-LOAM\cite{shan2018lego} and
A-LOAM\cite{qin2019aloam}. {The A-LOAM and LeGO-LOAM algorithms were obtained from their open-source code available on Github.} 
{The attempt to compare the open-source Intensity-SLAM~\cite{wang2021intensity} was unsuccessful as we could not run it.}
Our intensity based SLAM system shows competitive performance in different
environments, including indoor, outdoor, and some extreme scenarios,
such as long corridors. Most other LiDAR SLAM systems fail in such
extreme environment with fewer edge features.
We first tested our method with a public dataset provided by~Shan et
al.\cite{shan2021robust} which was collected by Ouster OS1-128 LiDAR. LIO-SAM's
trajectory\cite{liosam2020shan} was treated as the ground truth since it is
estimated based on LiDAR, 9-axis IMU and GPS, which is much more accurate than
LiDAR only SLAM. In Fig.~\ref{fig:trajtories_outdoor}, we present the
trajectories of our method, LeGO-LOAM, and A-LOAM in various terrains, while
Fig.~\ref{fig:trajtories_outdoor_ape} shows the corresponding Absolute Position
Error (APE) of Fig.~\ref{fig:trajtories_outdoor}. 
From Fig.~\ref{fig:trajtories_outdoor_ape}, we can say our method has a significantly lower (T-test\cite{kim2015t}, p\textless 1e-5) APE than others, except for the scenario in Fig.~\ref{fig:traj_indoor1_ape_box} where our result is not significant (T-test, p=0.857 compared with A-LOAM). 
The results in Fig.~\ref{fig:trajtories_outdoor} and Fig.~\ref{fig:trajtories_outdoor_ape} 
prove our proposed approach achieves competitive results compared to both A-LOAM and
LeGO-LOAM, especially in Fig.~\ref{fig:traj_jackal3} where the trajectory is
close to the ground truth trajectory in the end, while others drift a lot. {The effectiveness of LeGO-LOAM is limited to level terrains as it relies on a ground plane constraint. It becomes challenging to extract the ground plane information in uneven terrains, thereby hindering its performance in such environments.}

\begin{table}[]
        \caption[]{Time consumption of different algorithms (ms)}
        \label{tab:time_consumption}
        \centering
        \scalebox{0.8}{
        \begin{tabular}{llll}
        \hline
                                & \quad A-LOAM                & LeGO-LOAM & \qquad Ours \\ \hline
        Features extration      & \, 6.33 $\pm$ 1.99        &    10.78 $\pm$ 4.36         &   10.70 $\pm$ 1.50         \\
        Scan registration       & 20.04 $\pm$ 4.06      &  \, 2.53 $\pm$ 3.88          &  \, 3.15 $\pm$ 2.07         \\
        Map optimization        & 10.63 $\pm$ 2.61      &  21.61 $\pm$ 7.55           &   29.02 $\pm$ 6.23        \\
        PGO                     &  \qquad   N/A                &   23.48 $\pm$ 7.81         &  \, 1.73 $\pm$ 10.51   \\ \hline
        \end{tabular}}
\end{table}

We also test our algorithm indoors with a Spot robot (from Boston Dynamics)
equipped with Ouster Os0-64 LiDAR (Fig.~\ref{fig:spot_equipment}). 
The scene of this experiment mainly contains the same long corridor as Fig.~\ref{fig:experiment_env_robot_environment}. In this scenario, we ran different algorithms for testing the ability of localization and map building in real world. Both Fig.~\ref{fig:spot_corridor1_map} and Fig.~\ref{fig:spot_corridor2_map} are the maps generated by our algorithm. Fig.~\ref{fig:spot_corridor1_traj_xy} and Fig.~\ref{fig:spot_corridor2_traj} then show the trajectories of differnt algorithms in the corresponding environments. From the trajectories, we can see that LeGO-LOAM drifts a lot. A-LOAM's and our trajectories are almost the same. Due to the indoor environments, we can hardly collect the ground truth trajectories with RTK. So we try to analyze the map details (Fig.~\ref{fig:diff_aloamandour}) to evaluate the algorithm strengths and weaknesses. Fig.~\ref{fig:spot_corridor1_intensity_map_details} shows the front view of Fig.~\ref{fig:spot_corridor1_map}, and Fig.~\ref{fig:spot_corridor1_aloam_map_details} shows the front view of the same scenario generated by A-LOAM. We can see that our method can smoothly connect the start point at the end of the trip, but A-LOAM's map is disconnected. At this point, we can say that our method is more reliable in such an extreme environment.
In addition, we analyzed the time consumption of different SLAM algorithms on the Intel processor with the same data collected by Os0-64 LiDAR. Table~\ref{tab:time_consumption} shows that our intensity-based front-end is able to calculate the odometry within 15 ms and our method is efficient enough to meet the real-time requirements of 10 Hz LiDAR.

\section{CONCLUSIONS}
In this paper, we propose a novel intensity-based pure LiDAR SLAM method. We first proposed a novel lightweight intensity-based odometry method, which directly match 3D features point extracted from the intensity images. Then we propose a novel map optimization method, which jointly optimizes the LiDAR BA and point-to-plane residuals. Finally, we propose a novel intensity-based pose graph optimization method, which can optimize the pose graph based on the intensity image. We tested our method in both outdoor and indoor environments. The results proved that our method can achieve competitive results compared with other popular pure LiDAR SLAM methods. In the future, we will further improve our method by using more advanced feature extraction methods and loop closure detection methods. We will also test our method in more challenging environments.


\bibliographystyle{IEEEtran}
\bibliography{IEEEabrv,root}

\end{document}